%% file: asyndgan.tex
\documentclass[10pt,journal,compsoc]{IEEEtran}

\ifCLASSOPTIONcompsoc
  \usepackage[nocompress]{cite}
\else
  \usepackage{cite}
\fi
\ifCLASSINFOpdf
  \usepackage[pdftex]{graphicx}
\else
\fi
\usepackage{amssymb}
\usepackage{amsmath}
\usepackage{textcomp}

\usepackage{amsthm}
\newtheorem{thm}{Theorem}
\newtheorem*{remark}{Remark}

\newtheorem{lemma}{Lemma}

\theoremstyle{definition}

\usepackage{algorithmic}

\newsavebox{\ieeealgbox}

\usepackage{booktabs}
\usepackage{hyperref}
\begin{document}
%
\title{Multi-modal AsynDGAN: Learn From Distributed Medical Image Data without Sharing Private Information}
%
%
%
%

\author{Qi Chang,
        Zhennan Yan,
        Lohendran Baskaran,
        Hui Qu,
        Yikai Zhang,
        Tong Zhang,~\IEEEmembership{Fellow,~IEEE},
        Shaoting Zhang,
        and Dimitris N. Metaxas,~\IEEEmembership{Fellow,~IEEE}
\IEEEcompsocitemizethanks{\IEEEcompsocthanksitem Q. Chang, H. Qu, Y. Zhang, D.N. Metaxas are with the Department of Computer Science, Rutgers University, Piscataway, NJ, USA.\protect\\
E-mail: {qc58,dnm}@cs.rutgers.edu
\IEEEcompsocthanksitem Z. Yan is with SenseBrain Technology Limited LLC, Princeton, NJ, USA. E-mail: zhennanyan@sensebrain.site
\IEEEcompsocthanksitem L. Baskaran is with Dalio Institute of Cardiovascular Imaging, Weill Cornell Medicine, New York, NY, USA, and also with the Department of Cardiovascular Medicine, National Heart Centre, Singapore. E-mail: lohendran.baskaran@gmail.com
\IEEEcompsocthanksitem T. Zhang is with Hong Kong University of Science and Technology, Hong Kong, China. E-mail: tongzhang@tongzhang-ml.org
\IEEEcompsocthanksitem S. Zhang is with SenseTime Research, Shanghai, China. E-mail: zhangshaoting@sensetime.com}
\thanks{Manuscript received October 3, 2020; 
}}

%
%

\markboth{IEEE TRANSACTIONS ON PATTERN ANALYSIS AND MACHINE INTELLIGENCE}%
{Chang \MakeLowercase{\textit{et al.}}: Multi-modal AsynDGAN: Learn From Distributed Medical Image Data without Sharing Private Information}
%



\IEEEtitleabstractindextext{%
\begin{abstract}
As deep learning technologies advance, increasingly more data is necessary to generate general and robust models for various tasks. In the medical domain, however, large-scale and multi-parties data training and analyses are infeasible due to the privacy and data security concerns.
In this paper, we propose an extendable and elastic learning framework to preserve privacy and security while enabling collaborative learning with efficient communication. The proposed framework is named distributed Asynchronized Discriminator Generative Adversarial Networks (AsynDGAN), which consists of a centralized generator and multiple distributed discriminators.
The advantages of our proposed framework are five-fold: 1) the central generator could learn the real data distribution from multiple datasets implicitly without sharing the image data; 2) the framework is applicable for single-modality or multi-modality data; 3) the learned generator can be used to synthesize samples for down-stream learning tasks to achieve close-to-real performance as using actual samples collected from multiple data centers; 4) the synthetic samples can also be used to augment data or complete missing modalities for one single data center;
5) the learning process is more efficient and requires lower bandwidth than other distributed deep learning methods.
\end{abstract}

}

\maketitle

\IEEEdisplaynontitleabstractindextext

%
\IEEEpeerreviewmaketitle

\IEEEraisesectionheading{\section{Introduction}\label{sec:introduction}}


\IEEEPARstart{I}{n} recent years, machine learning techniques, especially deep learning~\cite{lecun2015deep}, have been adopted in a variety of medical image analysis tasks ~\cite{litjens2017survey,shen2017deep} with superior performance.
It's widely known that a sufficient amount of data samples is necessary for training a successful machine learning algorithm~\cite{domingos2012few}. Deep learning, which usually adopts a model with millions or even billions of parameters, requires even more training data samples to overcome the overfitting issue.

However, in the medical domain, acquiring enough medical images for training a deep learning model is not an easy task due to the privacy and security issues. 
On the one hand, multiple levels of regulations, such as HIPAA~\cite{annas2003hipaa,gostin2009beyond}, EU GDPR~\cite{regulation2018general}, and the approval process for the Institutional Review Board (IRB)~\cite{bankert2006institutional}, protect the patients' sensitive data from malicious copy or even tampering  of medical records~\cite{mirsky2019ct}. 
On the other hand, like a double-edged sword, these regulations cause difficulty for collaborations between institutions or countries. For example, many hospitals and research institutions are wary of cloud and distributed platforms, and prefer to use their own local server. 
It is even harder when the international collaboration is needed: America, European Union and many other countries do not allow patient information leave their country~\cite{seddon2013cloud,kerikmae2017challenges}. 

Another realistic challenge when collaborating from various entities is the misaligned modalities of datasets. Currently, multi-modal imaging data has been used in many clinical studies, for example, the tumor segmentation~\cite{menze2014multimodal,guo2019deep}, brain segmentation~\cite{ivsgum2015evaluation,wang2019benchmark}, and intervertebral disc localization and segmentation ~\cite{gao2019multi}. It is because methods fusing multi-information from multi-modality data commonly leads to improved performance than using a single modality~\cite{zhou2019review}. However, due to the different clinical protocols \cite{brown2018using,ellingson2015consensus} or various practical reasons across hospitals and countries, gathering all modalities among all institutes is not a trivial task.

We propose a distributed learning framework named  Distributed Asynchronized Discriminator GAN (AsynDGAN) that enable the secured collaboration across different medical entities and at the same time leverage the misalignment of modalities. In our designed framework, the centralized generator is used to learn the joint distribution of multiple data sets stored in different entities. Each entity has a discriminator, which is learned to classify the local real data and the synthetic data generated by the generator. This framework ensures the privacy and security of original image data without sharing with each data center or the central generator.
Through adversarial learning, a faithful estimation of the overall data distribution could be learned by the generator.
As a result, the generator can act as an image provider to synthesize data for some specific down-stream tasks, for example, image segmentation.
Learning from synthetic images has several advantages:

\textbf{Privacy mechanism}:
During the adversarial learning between generator and discriminators, only synthetic images, data annotations (e.g. mask of brain tumor for segmentation task) and losses are transmitted. AsynDGAN bridges the most obvious and gaping security issues in distributed machine learning by leaving the training data at its source.
In addition, data anonymization techniques could be avoided: achieving $100\%$ success rate of DICOM de-identification in a large dataset is not a trivial task due to the flexibility of DICOM data and various requirements for a specific study~\cite{aryanto2015free}.
Our mechanism avoids the sharing of raw image data, thus ensuring privacy and security more effectively.

\textbf{Synthetic data sharing}: 
Since synthetic images do not contain sensitive information, such an aggregation and redistribution system can build a publicly accessible and faithful medical database. The inexhaustible database can benefit researchers, practitioners and boost the development of medical intelligence.

\textbf{Adaptivity to architecture updates}: Deep learning techniques evolve so rapidly in recent years, 
that we could reasonably infer that the recently well-trained model may be outdated or underperform in the future as new techniques, such as network module, loss function or optimizer are invented. Due to different regulations and data usage agreements, the data may not be always accessible. 
Thus we may not embrace the new architectures to achieve higher performance when the dataset is off-lined.
Instead of training a task-specific model, our proposed method can learn a data generator to produce synthetic images for future use, without worrying about the loss of the proprietary data.

\textbf{Generation of multi-modality data}:
In tumor detection and brain disease diagnosis, multi-modality data are commonly used. Our framework is not only applicable to learn the joint distribution of single-modality imaging data, but also able to capture the overall distribution of multi-modality imaging data. Considering multi-modality data of the same region of interest (ROI) share the same content information with different appearance patterns, the central generator can learn to produce multiple modalities at the same time. Such a generator can be utilized for data augmentation or missing modality completion.

\textbf{Missing modality completion across data centers}:
In real world scenarios, the imaging protocols in different medical centers are not identical due to various technical and practical reasons. For example, selection of the appropriate protocol depends on the indication for the exam and the patient history~\cite{brown2018using,ellingson2015consensus} etc. For clinical studies involving multiple modalities, it is harder to collect a good size of database from different institutes since multi-modality data are usually incomplete. Unlike existing methods that generate the missing modality from another available modality~\cite{cai2018deep}, our proposed framework could tackle missing modalities by learning to generate complete sets of image modalities from different data centers. Note that our collaborative learning is based on the assumption that the missing modality of one data center can be found in other data centers.

Our proposed approach is a substantial extension from our previous work~\cite{chang2020synthetic}. In this paper, we further extend our method to support multi-modality data learning. It is worth noting that the proposed framework can handle the situation when the data centers have missing modalities and the learned generator can be used for missing modality completion. We also improve the experimental settings by using data centers with various size of dataset. The results show that the proposed method can successfully learn to synthesize single-modality or multi-modality data across data centers without direct access and achieve performance improvement by missing modality completion or synthetic data augmentation.

In summary, our contributions are five-fold: (1) a distributed asynchronized discriminator GAN (AsynDGAN) is proposed to learn the joint distribution of real images from different data centers without sharing user's private data; (2) the proposed framework can be applicable for multiple modalities and various distribution dataset; (3) the learned generator can be used to synthesize training samples for down-stream learning tasks to achieve close-to-real performance as using actual samples collected from multiple data centers; (4) the synthetic samples can also be used to augment data or complete missing modalities for one single data center; (5) the learning process is more efficient and requires lower bandwidth than other federated learning methods.

\section{Related Work}
\label{sec:related-work}
\input{sec_related_work}

\section{Method}
\label{sec:method}

\subsection{Network architecture}

\begin{figure*}[ht]
\begin{center}
\includegraphics[width=0.9\linewidth,height=5.5cm]{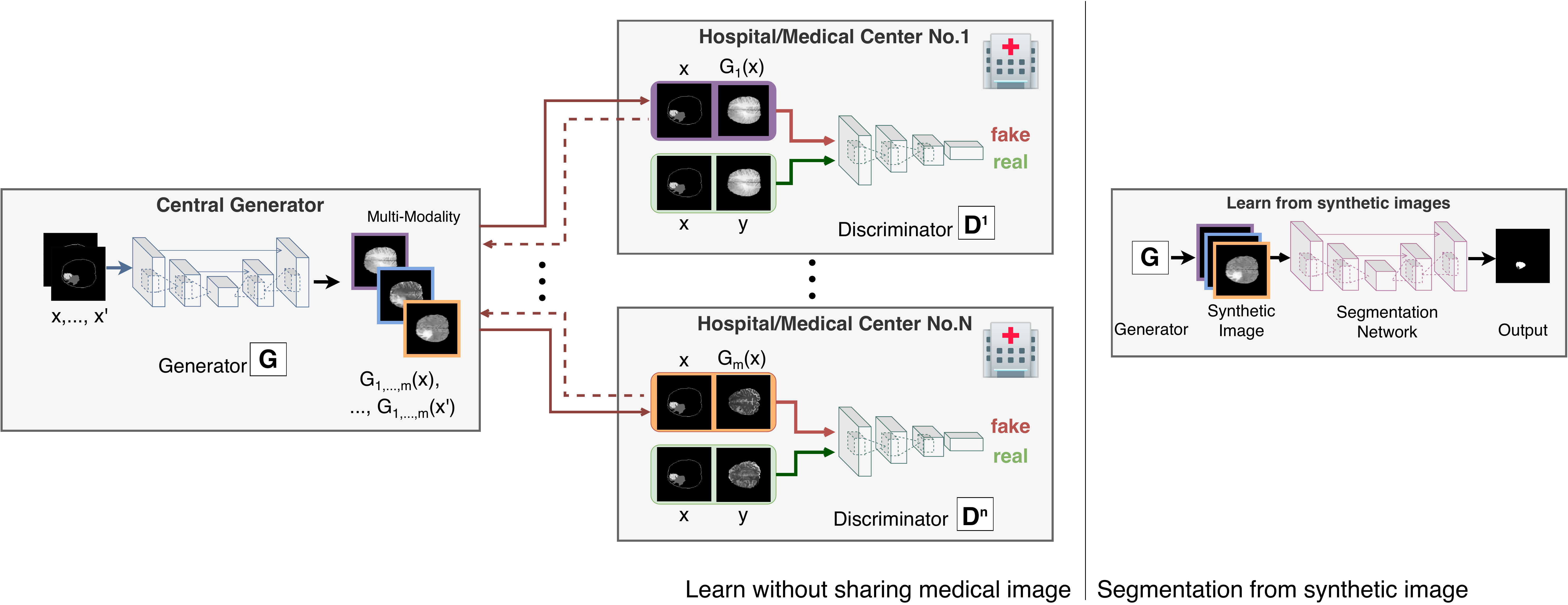}
\end{center}
\caption{The overall structure of multi-modality AsynDGAN. It contains two parts, a central generator $G$ and multiple distributed discriminators $D^1, D^2, \cdots, D^n$ in each medical entity. $G$ takes a task-specific input (segmentation masks in our experiments) and outputs multi-modality synthetic images. Each discriminator learns to differentiate between the real images of current medical entity and synthetic images from $G$. The well-trained $G$ is then used as an image provider to train a task-specific model (segmentation in our experiments).}
\label{arch1}
\end{figure*}

Our proposed AsynDGAN is comprised of one central generator and multiple distributed discriminators located in different medical entities.
An overview of the proposed architecture is shown in Figure~\ref{arch1}.
The central generator, denoted as $G$, takes task-specific inputs (e.g. segmentation masks in our use case) and generates synthetic images to fool the discriminators. Let $N$ denote the number of medical entities that involve in the learning framework, and $S_j=\{(x_i^j,y_i^j):i=1,...,s^j\}$ denote the local private dataset at the $j$-th entity, where $x$ is an auxiliary variable representing a class label or a mask, and $y$ is the corresponding real image data. The local discriminators, denoted as $D_j$, $j\in\{1,...,N\}$, learn to differentiate between the local real images $y_i^j$ and the synthetic images $\hat{y}_i^j=G(x_i^j)$ generated from $G$ based on $x_i^j$.
Our architecture ensures that $D_j$ deployed in the $j$-th medical entity only has the access to its local dataset, while not sharing any real image data outside the entity. During the learning process, Only synthetic images, masks, and losses are transferred between the central generator and the discriminators. Such design naturally complies with privacy regularizations and keeps the patients' sensitive data safe. 

After the adversarial learning, the generator will learn the joint distribution from different datasets that belong to different medical entities. Then, it can be used as an image provider to generate training samples for some down-stream tasks. Assuming the distribution of synthetic images, $p_{\hat{y}}$, is same or similar to that of the real images, $p_{data}$, we can generate one unified large dataset which approximately equals to the union of all the datasets in medical entities. In this way, all private image data from each entity are utilized without sharing. In order to evaluate the synthetic images, we use the generated samples in segmentation tasks to illustrate the effectiveness of proposed AsynDGAN. Details about $G$ and $D_j$'s designed for segmentation tasks are described below.

\subsubsection{Central generator}
For segmentation tasks, the central generator is designed to generate images based on input masks, so that the synthetic image and corresponding mask can be used as a pair to train a segmentation model. Here, an encoder-decoder network is adopted for $G$. It consists of nine residual blocks~\cite{he2016resnet}, two stride-2 convolutions for downsampling and two transposed convolutions for upsampling. All non-residual convolutional layers are followed by batch normalization~\cite{ioffe2015batch} and the ReLU activation. All convolutional layers use $3\times3$ kernels except the first and last layers that use $7\times7$ kernels.

\subsubsection{Distributed discriminators}
In our framework, each discriminator has the same structure as that in PatchGAN~\cite{pix2pix2017}. The discriminator classifies a small image patch as real or fake. Such architecture assumes patch-wise independence of pixels in a Markov random field fashion \cite{li2016precomputed,isola2017image}, and the patch is large enough to capture the difference in geometrical structures such as background and tumors. 

\subsection{Objective of AsynDGAN}
The AsynDGAN is based on the conditional GAN~\cite{mirza2014conditiongan}. The objective function is:
\begin{equation}
\begin{aligned}
&\min\limits_{G}\max\limits_{D_1:D_N}V(D_{1:N},G) \\
&= \sum\limits_{j\in [N]} \pi_j \{\mathbb{E}_{x\sim s_j(x)}[\mathbb{E}_{y\sim p_{data}(y,x)} \log D_j(y,x) \\
&+\mathbb{E}_{\hat{y}\sim p_{\hat{y}}(\hat{y},x)} \log(1-D_j(\hat{y},x))]\}
\end{aligned}
\label{eq:cgan}
\end{equation}
The goal of $D_j$ is to maximize Eq.~\ref{eq:cgan}, while $G$ minimizes it. In this way, the learned $G(x)$ with maximized $D(G(x))$ can approximate the real data distribution $p_{data}(y|x)$ and $D$ cannot tell `fake' data from real. $x$ follows a distribution $s(x)$. 
In this paper, We assume that $s(x)=\sum\limits_{j\in[N]} \pi_js_j(x)$, where $s_j(x)$ is marginal distribution of $j$-th dataset  and $\pi_j$ is the prior distribution. In the experiment, we set $s_j(x)$ be a uniform distribution and $\pi_j \propto |S_j|$. For each sub-distribution, there is a corresponding discriminator $D_j$ which only receives data generated from prior $s_j(x)$.
Similar to previous works~\cite{mathieu2015deep,pix2pix2017}, instead of providing Gaussian noise $z$ as an input to the generator, we provide the noise only in the form of dropout, which applied to several layers of the generator $G$ at both training and inference.
The optimization process is formulated as Algorithm~\ref{algo1}.
The losses of $D_j$ and $G$ are defined in Eq.~\ref{eq:lossD} and Eq.~\ref{eq:lossG}, respectively.
\begin{equation}
    L_{D_j} = \frac{1}{m} \sum_{i=1}^m \left[
			- \log D_j(y_i^j,x_i)
			- \log (1-D_j(\hat{y}_i^j,x_i))
			\right],
\label{eq:lossD}
\end{equation}
\begin{equation}
\begin{aligned}
    L_G = \frac{1}{Nm} \sum_{j=1}^N\pi_j\sum_{i=1}^m 
			[ &\log (1-D_j(\hat{y}_i^j,x_i))\\
			&+ \lambda_1 L_1(y_i^j, \hat{y}_i^j) + \lambda_2 L_{P}(y_i^j, \hat{y}_i^j)].
\end{aligned}
\label{eq:lossG}
\end{equation}
where $m$ is the minibatch size. The $L_G$ contains perceptual loss ($L_{P}$)~\cite{Johnson2016Perceptual} and $L_1$ loss besides of the adversarial loss.




\begin{figure}
	\caption{\small Training algorithm of AsynDGAN.
	}
	\begin{algorithmic}\label{algo1}
		\FOR{number of total training iterations}
		\FOR{number of interations to train discriminator}
		\FOR{each node $j \in [N]$}
		\STATE{-- Sample  minibatch of of $m$ auxiliary variables $\{x^j_1,...,x^j_m\}$ from $s_j(x)$ and send to generator $G$.}
		\STATE{-- Generate $m$ fake data from generator $G$, $\{\hat{y}^j_1,...,\hat{y}^j_m\}$, where $\hat{y}^j_k=G(x^j_k)$, and send to node $j$.}
		\STATE{-- Update the discriminator $D_j$ by ascending its stochastic gradient:
			$\nabla_{\theta_{D_j}} L_{D_j}$.
			}
		\ENDFOR
		\ENDFOR
		\FOR{each node $j \in [N]$}
		\STATE{-- Sample  minibatch of $m$ auxiliary variables $\{x^j_1,...,x^j_m\}$ from $s_j(x)$ and send to generator $G$.}
		\STATE{-- Generate corresponding $m$ fake data from generator $G$, $\{\hat{y}^j_1,...,\hat{y}^j_m\}$ and send to node $j$.}
		\STATE{-- Discriminator $D_j$ passes error to generator $G$.}
		\ENDFOR
		\STATE{-- Update $G$ by descending its stochastic gradient:
			$\nabla_{\theta_G} L_G.$}
		\ENDFOR
		\\The gradient-based updates can use any standard gradient-based learning rule. We used momentum in our experiments.
	\end{algorithmic}
\end{figure}

\subsection{Analysis: AsynDGAN learns the correct distribution}

In this section, we present a theoretical analysis of AsynDGAN and discuss the implications of the results. We first begin with a technical lemma describing the optimal strategy of the discriminator.

\begin{lemma}\label{lem1}
	When generator $G$ is fixed,  the optimal discriminator $D_j(y|x)$ is :\\
	\begin{equation}
	D_j(y|x)=\frac{p(y|x)}{p(y|x)+q(y|x)}
	\end{equation}
	where $p(y|x)$ is the real data distribution and $q(y|x)$ is the synthetic data distribution.
\end{lemma}

Suppose in each training step the discriminator achieves its maxima criterion in Lemma \ref{lem1}, the objective of the generator becomes:
\begin{equation*}
\begin{aligned}
\min\limits_{G}V(G)&= \mathbb{E}_{x}\mathbb{E}_{y\sim p_{data}(y|x)} [\log D(y,x)] \\
&+\mathbb{E}_{\hat{y}\sim p_{\hat{y}}(\hat{y}|x)} [\log(1-D(\hat{y},x))]\\
&=\sum_{j\in[N]} \pi_j\int\limits_{x} s_j(x)\int\limits_{y} p(y|x)\log\frac{p(y|x)}{p(y|x)+q(y|x)}\\
&+q(y|x)\log\frac{q(y|x)}{p(y|x)+q(y|x)} dydx\\
\end{aligned}
\end{equation*}

Assuming in each step, the discriminator always performs optimally, we show indeed the generative distribution $G$ seeks to minimize the loss by approximating the underlying distribution of data.
\begin{thm}
	Suppose the discriminators $D_{1... N}$ always behave optimally (denoted as $D^*_{1 ... N}$), the loss function of generator is global optimal iff $q(y,x)=p(y,x)$ where the optimal value of $V(G,D^*_{1... N})$ is $-\log 4$. 
\end{thm}

\begin{remark}
	While analysis of AsynDGAN loss shares similar spirit with~\cite{goodfellow2014generative}, it has different implications. In the distributed learning setting, data from different nodes are often dissimilar. Consider the case where $\Omega(s_j(x)) \cap \Omega(s_k(y)) =\emptyset, \text{for } k \neq j$, the information for $p(y|x), y\in \Omega(s_j(x))$ will be missing if we lose the $j$-th node. The behavior of trained generative model is unpredictable when receiving auxiliary variables from unobserved  distribution $s_j(x)$.
	The AsynDGAN framework provides a solution for unifying different datasets by collaborating multiple discriminators.
\end{remark}

\subsection{Learn from multi-modality data}

For use cases of multi-modality data, the local datasets have multi-modality image  $\pmb{y}_{i}^j=(y_{i,1}^j,...,y_{i,c}^j)$ associated with each $x_i^j$.
A simple way of handling the multi-modality image in our framework would be treating the $c$ modalities of one sample as a $c$-channel image, and thus the only change needed is the number of channels of input layer of $D$ and output layer of $G$. In this setting, the learning task of $D$ could be easier and converge very fast since different modalities have different contrast patterns and there are more information can be used to differentiate the real and the 'fake' data. However, the task of $G$ may become more difficult to learn. It is because on one hand the $G$ needs to learn more complex data distribution to generate multiple modalities with different contrasts, on the other hand the easily-learned $D$ may learn some trivial discriminative features and thus cannot provide helpful feedback to $G$ to guide its learning.

In order to balance the task difficulty of the $G$ and $D$'s, we extend our framework by deploying multiple discriminators at each entity. In one data center, every single modality has its own discriminator and the $G$ receives losses from the multiple $D$'s for a multi-modality data sample. In this way, each $D$ can focus on learning discriminative features for one specific modality and provide more meaningful feedback to $G$. The objective function can be extended from Eq.~\ref{eq:cgan} as:
\begin{equation}
\begin{aligned}
&\min\limits_{G}\max\limits_{D_{1:N}^{1:c}}V(D_{1:N}^{1:c},G) \\
&= \sum\limits_{j\in [N]} \pi_j \{\mathbb{E}_{x\sim s_j(x)} \sum\limits_{k=1}^c [ \mathbb{E}_{y_k\sim p_{data}(y_k|x)} \log D_{j,k}(y_k|x) \\
&+\mathbb{E}_{\hat{y_k}\sim p_{\hat{y}}(\hat{y_k}|x)} \log(1-D_{j,k}(\hat{y_k}|x))]\},
\end{aligned}
\label{eq:mdcgan}
\end{equation}
where $D_{j,k}$ represents the discriminator for the $k$-th modality at the center $j$.

Besides, another advantage of the proposed multi-modality framework is that it enables learning from missing modality data. Let $C_{j}$ denote the set of index of available modality for center $j$, if data center $j$ misses the $s$-th modality for example, then $C_{j}=\{1,...,s-1,s+1,...,c\}$. In this case, center $j$ only need to deploy $c-1$ discriminators during the learning. The learning process has no difference except that it only collects losses of available discriminators for $C_j$ to update the $G$ and only use a subset of the synthetic images $\{\hat{y}_k^j|k\in C_{j}\}$ to update the corresponding $\{D_{j,k}|k\in C_{j}\}$ in center $j$. 
Because the discriminators for different modalities in different entities are all independent, the $G$ can still learn to generate all modalities, assuming that the missing modality in one center is available in some other data centers. 
The loss function of $D$ is the same, while the loss function of $G$ can be further extended as the following:
\begin{equation}
\begin{aligned}
    L_G = \frac{1}{Nm} \sum_{j=1}^N & \pi_j\sum_{i=1}^m \sum_{k\in C_j}
			[\log (1-D_{j,k}(\hat{y}_{i,k}^j,x_i)) \\
			&+ \lambda_1 L_1(y_{i,k}^j, \hat{y}_{i,k}^j) + \lambda_2 L_{P}(y_{i,k}^j,\hat{y}_{i,k}^j)].
\end{aligned}
\label{eq:lossG_miss}
\end{equation}


After training, the learned $G$ can act as a synthetic image provider to generate multi-modality images from the conditional variable, a mask image. As a result, it can also be used for missing modality completion. For instance, if a data center has data $(y_{1},...,y_{s-1},y_{s+1},y_{c})$ with the $s$-th modality missing and the corresponding mask image $x$, we can use the synthetic image $\hat{y}_s=G_s(x)$ as a substitute.

\begin{figure}[t]
	\begin{center}
		\includegraphics[width=0.75\linewidth]{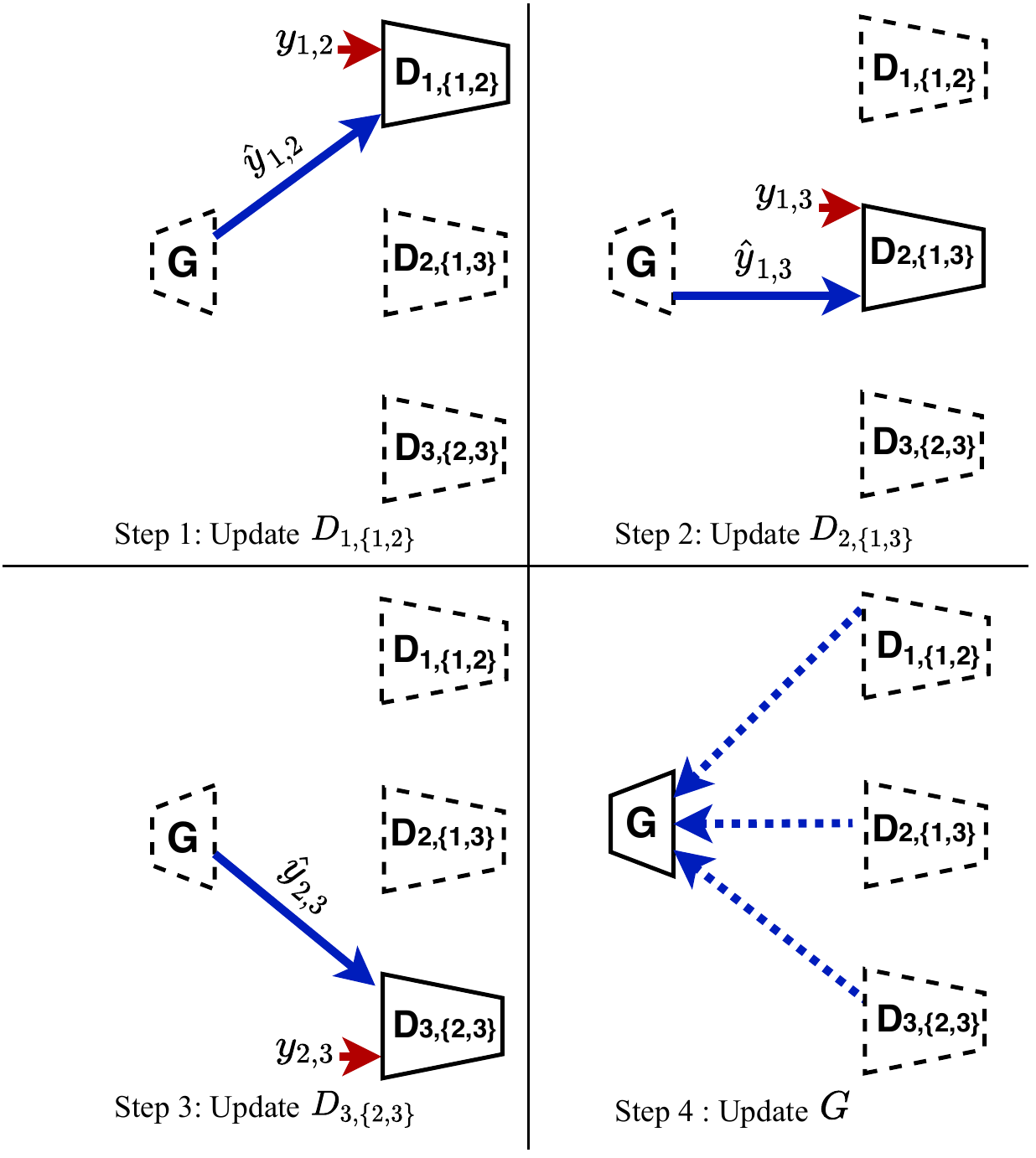}
	\end{center}
	\caption{The optimization process of AsynDGAN when learning from distributed multi-modality datasets. Each dataset may have different modalities available. For example, $D_{2,\{1,3\}}$ is a dataset with the second modality missing. The solid arrows show the forward pass of synthetic images to update the discriminators, and the dotted arrows show the backward flow to update the generator. The dotted blocks indicate the fixed components while the solid block is updating.}
	\label{workflow}
\end{figure}

Our approach is different from the existing methods that predict the target modality from another modality~\cite{cai2018deep,Zhang_2018_CVPR} in the sense that it can generate multiple modalities to handle randomly missing modality problem, and thus does not require a specific model for specific modality pair for the input and output. 

\section{Experiments}
\label{sec:exp}
\input{sec_exp}

\section{Conclusion}
\label{sec:conclude}
In this work, we proposed a distributed GAN based learning framework as a solution to the privacy restriction problem in sharing medical data among different institutes for collaborative studies. Our framework includes a central generator and multiple distributed discriminators. 
We stress that the learning framework is general and could be collaborated with variants of GAN loss, e.g. Wasserstein distance and classical regression loss~\cite{arjovsky2017wasserstein,mao2017least}, and different network architectures. 

There are several advantages of the proposed framework. Firstly, the framework can be a good solution to the privacy issue in collaborative studies that involve multiple institutes. Because the generator can learn the joint data distribution of multiple datasets in different health entities without direct accessing or storing patients' private image data. The learning process is more efficient than other federated learning approaches in terms of communication between server and distributed nodes.

Secondly, the well-trained generator can be used as an image provider to create a large database of synthetic images for training task-specific models. Therefore, iterative refinement of the task-specific models can be done locally and does not require the access to any private dataset or maintenance of a large database.

Last but not least, our framework is easily extensible.
We have demonstrated its application for both single- and multi-modality image generation in this study. It can also handle the missing modality problem. 
The performance is evaluated by a tumor segmentation task.
In our experiments, the segmentation model trained solely by synthetic data has a comparable performance with the model trained by all real data, and outperforms models trained by local data in each medical entity. In addition, the synthetic data can also be used as data augmentation to increase the size and variation of the training database to boost performance.
Furthermore, the learned generator can be used for missing modality completion. The generated missing modality can help in improving model performance together with other available real modalities.

In the future work, we will investigate further improvement of the conditional information of the generator to obtain better control of its generated context besides the masked region (e.g. the brain tissues outside the tumor region in this use-case).


%



\ifCLASSOPTIONcompsoc
  \section*{Acknowledgments}
\else
  \section*{Acknowledgment}
\fi

This work was supported by the National Science Foundation [grant numbers NSF-1733843, NSF-1763523, NSF-1747778, NSF-1703883, NSF-1909038, NSF-1855759 and NSF-1855760]

\ifCLASSOPTIONcaptionsoff
  \newpage
\fi



\bibliographystyle{IEEEtran}
\bibliography{refs}

%

\begin{IEEEbiography}{Qi Chang}
Biography text here.
\end{IEEEbiography}

\begin{IEEEbiography}{Zhennan Yan}
Biography text here.
\end{IEEEbiography}

\begin{IEEEbiography}{Lohendran Baskaran}
Biography text here.
\end{IEEEbiography}

\begin{IEEEbiography}{Hui Qu}
Biography text here.
\end{IEEEbiography}

\begin{IEEEbiography}{Yikai Zhang}
Biography text here.
\end{IEEEbiography}

\begin{IEEEbiography}{Tong Zhang}
Biography text here.
\end{IEEEbiography}

\begin{IEEEbiography}{Shaoting Zhang}
Biography text here.
\end{IEEEbiography}

\begin{IEEEbiography}{Dimitris N. Metaxas}
Biography text here.
\end{IEEEbiography}






\end{document}


\section{AsynDGAN learns the correct distribution}
	In this section we present an analysis of AsynDGAN and discuss the implications of the results. We show that the AsynDGAN is able to aggregates multiple separated data set and learn generative distribution in an \emph{all important} fashion.
	We first begin with a technical lemma describing the optimal strategy of the discriminator.
	\begin{lemma}\label{lem1}
		When generator $G$ is fixed,  the optimal discriminator $D_j(y,x)$ is :\\
		\begin{equation}
		D_j(y,x)=\frac{p(y|x)}{p(y|x)+q(y|x)}
		\end{equation}
	where $p(y|x)$ is the real data distribution and $q(y|x)$ is the synthetic data distribution.
	\end{lemma}
	\textbf{Proof}:\\
	\begin{equation*}
	\begin{aligned}
	&\max\limits_{D}V(D)=\max\limits_{D_1...D_N}\sum \pi_j\int\limits_{x} s_j(x)\int\limits_{y} p(y|x)log D_j(y,x)+q(y|x)log(1-D_j(y,x)) dydx\\
	&\leq \sum \pi_j\int\limits_{x} s_j(x)\int\limits_{y} \max\limits_{D_j} \{p(y|x)log D_j(y,x)+q(y|x)log(1-D_j(y,x)) \}dydx
	\end{aligned}
	\end{equation*}
	by setting $D_j(y,x)=\frac{p(y|x)}{p(y|x)+q(y|x)}$ we can maximize each component in the integral thus make the inequality hold with equality.\qed
	
	Suppose in each training step the discriminator achieves its maxima criterion in Lemma \ref{lem1}, the loss function for the generator becomes:\\
	\begin{equation*}
	\begin{aligned}
	&\min\limits_{G}V(G)= \mathbb{E}_{x}\mathbb{E}_{y\sim p_{data}(y|x) }[logD(y,x)] +\mathbb{E}_{\hat{y}\sim p_{\hat{y}}(\hat{y}|x)} [log(1-D(\hat y,x))]\\
	&=\sum_{j\in[N]} \pi_j\int\limits_{x} s_j(x) \underbrace{\int\limits_{y} p(y|x)log\frac{p(y|x)}{p(y|x)+q(y|x)}+q(y|x)log\frac{q(y|x)}{p(y|x)+q(y|x)} dydx}_{\text {To be analyzed in Lemma \ref{lem2}}}
	\end{aligned}
	\end{equation*}
	
	\begin{lemma} \label{lem2}
		Let $a(y)$ and $b(y)$ be two probability distributions s.t. support  $\Omega(a)\subset \Omega(b)$, the loss function $L(a)=\int\limits_{y} a(y)log\frac{a(y)}{a(y)+b(y)}+b(y)log\frac{b(y)}{a(y)+b(y)} dy \geq  -\log4$. The equality holds iff $b(y)=a(y)$.
	\end{lemma}
	
	\textbf{Proof}:\\
	Let $\lambda$ be the Lagrangian multiplier. 
	\begin{equation}
	\begin{aligned}
	L(a,\lambda)=\int\limits_{y} a(y)log\frac{a(y)}{a(y)+b(y)}+b(y)log\frac{b(y)}{a(y)+a(y)} + \lambda a(y) \;\;dy -\lambda
	\end{aligned}
	\end{equation}
	By setting $\frac{\partial L}{\partial a} =0$ we have $log\frac{a(y)}{a(y)+b(y)}=\lambda$ holds for all $x$. The fact that $log\frac{a(y)}{a(y)+b(y)}$ is a constant enforces $a(y)=b(y)$. Plugging $a(y)=b(y)$ into loss function we have $L(a)_{|a(y)=b(y)}=-\log4$.  \\
	Now we are ready to prove our main result that AsynDGAN learns the correct distribution. Assuming in each step, the discriminator always perform optimally, we show indeed the generative distribution $G$ seeks to minimize the loss by approximating underlying generative distribution of data.
	\begin{thm}
		Suppose the discriminators $D_{1... N}$ always behaves optimally (denoted as $D^*_{1 ... N}$), the loss function of generator is global optimal iff $q(y,x)=p(y,x)$ where the optimal value of $V(G,D^*_{1... N})$ is $-log 4$. 
	\end{thm}
	
	\textbf{Proof}:\\
	\begin{equation*}
	\begin{aligned}
	&\min\limits_{\substack{ q(y,x)>0,\\\int\limits_{y} q(y,x)=s(x)}}\sum\limits_{j\in[N]} \pi_j \int\limits_{x}s_j(x)\int\limits_{y} p(y|x)log\frac{p(y|x)}{p(y|x)+q(y|x)}+q(y|x)log\frac{q(y|x)}{p(y|x)+q(y|x)} dydx\\
	&\geq\sum\limits_{j\in[N]} \pi_j \int\limits_{x}s_j(x) \min\limits_{\substack{ q(y|x)>0,\\\int\limits_{y} q(y|x)=1}}\int\limits_{y} p(y|x)log\frac{p(y|x)}{p(y|x)+q(y|x)}+q(y|x)log\frac{q(y|x)}{p(y|x)+q(y|x)} dydx\\
	\end{aligned}\\
	\end{equation*}
	By Lemma \ref{lem2}, the optimal condition for minimizing:\\
	\begin{equation*}
	\min\limits_{\substack{ q(y|x)>0,\\\int\limits_{y} q(y|x)=1}}\int\limits_{y} p(y|x)log\frac{p(y|x)}{p(y|x)+q(y|x)}+q(y|x)log\frac{q(y|x)}{p(y|x)+q(y|x)} dydx
	\end{equation*}
	is by setting $q(y|x)=p(y|x), \forall x$. Such choice of $q(y|x)$ makes the inequality holds as an equality. Meanwhile $q(y|x)=p(y|x)$ implies $q(y,x)=p(y,x)$ given the fact that $p,q$ has the same marginal distribution on $y$. By plugging in $q(y|x)=p(y|x)$ we can derive the optimal value of $V(G,D^*_{1... N})$ to be $-\log4$.
	
	
	{\small
		\bibliographystyle{ieee_fullname}
		\bibliography{refs}
	}

%% file: sec_related_work.tex
\subsection{Learning with data privacy}
\textbf{Federated Learning}: 
The federated learning (FL) enables collaborative learning among local nodes in the network without sharing data. This feature makes it an attractive approach recently due to the privacy concern~\cite{konevcny2016federated,hard2018federated,brisimi2018federated,huang2018loadaboost}. To learn a globally powerful model, FL only communicates model information (parameters, gradients) between server and local nodes without transferring or accessing users' data. This mechanism plugs the most obvious and gaping security issues in distributed machine learning by leaving the training data at its source. It protects the privacy of user-data in different ways of various situations, such as by using differential privacy~\cite{wagner2018technical,dwork2014algorithmic, abadi2016deep} and homomorphic encryption~\cite{gentry2009fully,gentry2009fully2}.
Similar to the Federated Learning, AsynDGAN can easily apply the above methods to prevent many malicious attacks like inference attack\cite{orekondy2018gradient,melis2019exploiting}.

The distributed stochastic gradient descent (SGD) can also be shared in a privacy protection fashion in \cite{mcmahan2016communication,agarwal2018cpsgd,jayaraman2018distributed}. 
However, communicating gradients of a complex deep learning model is not efficient, for example, learning a ResNet101~\cite{he2016resnet} network with more than 40 million parameters requires at least $170$ MB to transfer gradients for each node per iteration.
Even with some compression technique~\cite{agarwal2018cpsgd}, the communication cost is still significant.


\textbf{Split Learning}: 
The split learning (SL)~\cite{vepakomma2018split} separates shallow and deep layers in deep learning models. The central processor only maintains layers that are several blocks away from the local input, and only inter-layer information is transmitted from local to central. In this way, the privacy is guaranteed because the central processor has no direct access to data. It reduces the communication cost from model-dependent level to cut-layer-dependent layer while protecting data privacy. However, such method does not apply to neural networks with skip connections, e.g., ResNets~\cite{he2016resnet}.


\textbf{Distributed GANs}:
The Generative Adversarial Network (GAN)~\cite{goodfellow2014generative} is designed to estimate generative data distribution via adversarial learning between a generator $G$ and a discriminator $D$. \cite{hardy2019md} extended the standard GAN to a multi-discriminator GAN (MD-GAN) in a distributed fashion. The discriminators are located in a set of worker machines where local data are stored. MD-GAN swaps discriminator models between workers at each iteration and aims at training the same $D$ by using all local data.


Comparing to Federated Learning and Split Learning, the communication cost of AsynDGAN is much lower since it is independent of the model parameter dimension. Only auxiliary data (label and masks), generated `fake' data and losses are transferred between the central server and local nodes during learning. For a $256\times256$ size gray-scale image, communication cost per-iteration for each node is about $8$ mb with batch size $32$. In addition, adaptivity is an exclusive advantage of AsynDGAN framework since it can generate synthetic training data for future training use. 

Unlike the setting of MD-GAN, AsynDGAN~\cite{chang2020synthetic} learns the joint distribution from different datasets that belongs to different medical entities without parameter sharing between different discriminators, and thus is much more efficient in bandwidth consumption, especially when the number of data center increases significantly.




\subsection{Multi-modality medical image segmentation}

As deep learning and multi-modality medical imaging techniques advance, deep learning based multi-modality medical image segmentation is obtaining more and more attention in recent years~\cite{zhou2019review}. \cite{zhou2019review} has reviewed recent deep learning based multi-modality fusion methods for medical image segmentation.
Since missing modality is a common problem in practice, some studies have tried to address it to help disease diagnosis \cite{li2014deep,cai2018deep}. For example, \cite{cai2018deep} utilized a GAN architecture combined with a classification loss to generate the missing modality data.
In this work, our framework can not only learn to generate different modalities simultaneously from multiple private datasets, but also learn from data centers with missing modality data. We show that the learned generator can be used to obtain complete multi-modality training data by synthesizing the missing modality for image segmentation tasks.

%% file: sec_exp.tex
In this section, 
we first perform the experiments on a toy task to illustrate how AsynDGAN learns a mixed Gaussian distribution from different subsets. We then apply AsynDGAN on a real MRI dataset, BraTS2018, and evaluate in three different settings. In the first setting, we use homogeneous data centers with single-modal images. Each local dataset has the same size and roughly the same data distribution. In the second setting, we extend our work on generating multi-modal images from three heterogeneous data centers. In this case, each dataset has different size and different data distribution. Lastly, we explore the AsynDGAN's adaptability to the modality variances across data centers. For instance, centers may have different missing modalities. With different settings among the data centers and modalities, we could evaluate the performance of AsynDGAN towards a real-world scenario. Without loss of generality, we adopt the image segmentation as the down-stream task described in this paper.


\subsection{Evaluation on a synthetic task}

\begin{figure}
\centering
\begin{minipage}{\linewidth}
	\centering\includegraphics[width=\linewidth]{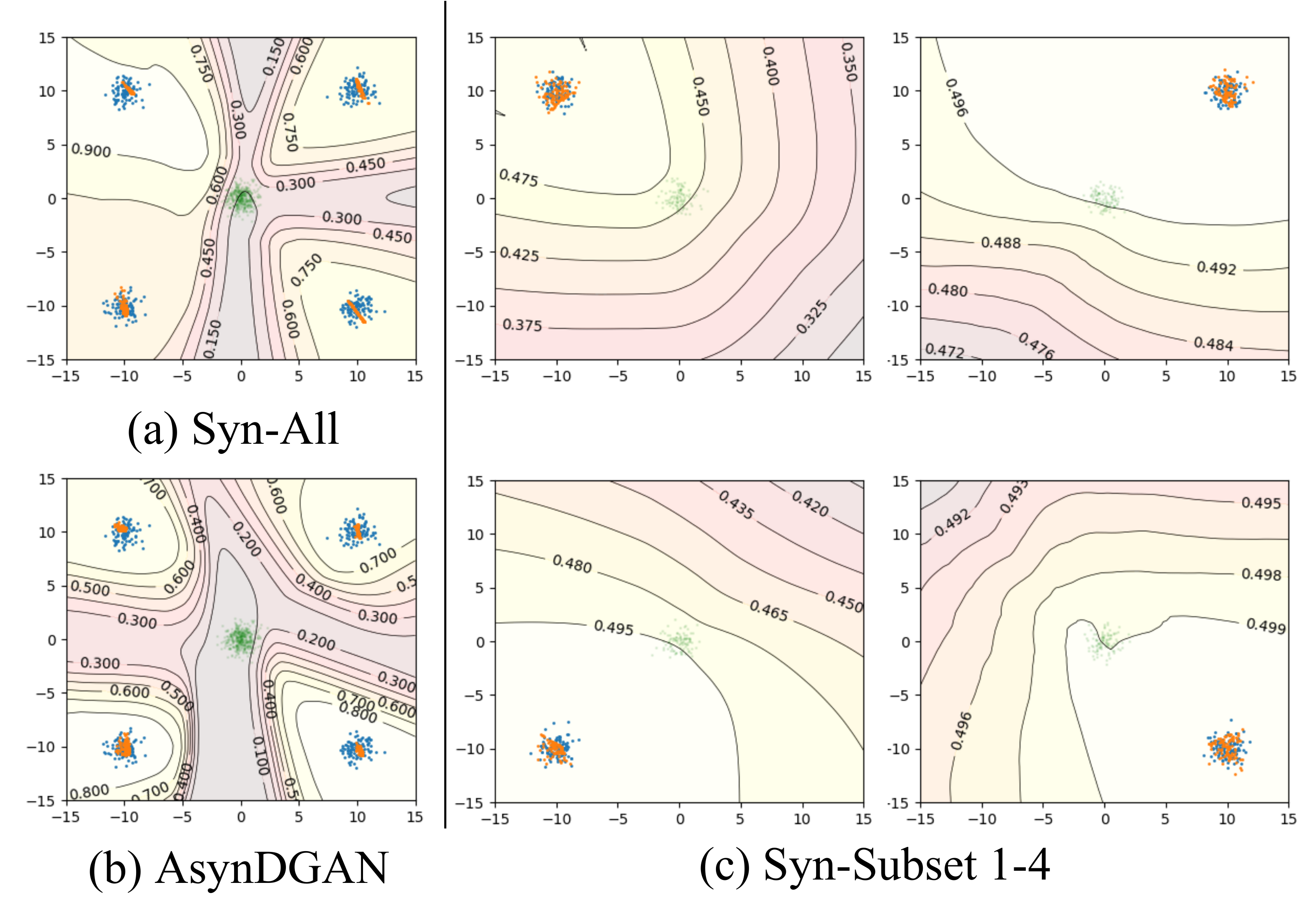}
\end{minipage}
\caption{Generated distributions of different methods. The blue points are data samples of 4 local datasets with different distribution. The green points are Gaussian noise (input) centered at (0, 0) and the orange points are synthetic samples generated by different methods.}
\label{fig:gaussian}
\end{figure}

We first evaluate our method on a toy task and show that the proposed synthetic learning framework can learn a mixture of Gaussian distribution from different subsets.
In this toy task, 4 local datasets contain 2-D data samples generated by 4 different multivariate Gaussian distributions centered at (10, 10), (10, -10), (-10, 10), (-10, -10), respectively. The covariance matrices are the same, i.e. $diag\{0.5, 0.5\}$. In Fig~\ref{fig:gaussian}, the data samples are shown as blue points. The central generator takes Gaussian noise, centered at $(0, 0)$ with covariance $diag\{0.5, 0.5\}$ (green points), as input and learns to transform them into points matching the mixture distribution (orange points).

The synthetic learning is conducted in $3$ settings: (1) Syn-All. Training a regular GAN by using all samples collected from the union of all the local datasets. (2) AsynDGAN. Training our AsynDGAN using all local subsets in a distributed fashion without direct access.
(3) Syn-Subset 1-4. Training 4 regular GAN's by only using samples in the corresponding local subset $n$, where $n\in\{1, 2, 3, 4\}$. 


The learned distributions are shown in Fig~\ref{fig:gaussian}. The GAN in Syn-All can learn the mixture distribution as expected since it can access all data samples from local subsets. If data sharing between subsets is prohibited, the local learning can only fit one of the Gaussians as shown in Fig~\ref{fig:gaussian}(c), while AsynDGAN is able to capture the joint distribution and thus has a comparable performance with the Syn-All.

\subsection{Real dataset and evaluation metrics}
\subsubsection{Dataset}


The BraTS2018 dataset comes from the Multimodal Brain Tumor Segmentation Challenge 2018~\cite{menze2014multimodal,bakas2017advancing,bakas2018identifying}. It contains multi-parametric magnetic resonance imaging (mpMRI) scans of low-grade glioma (LGG) and high-grade glioma (HGG) patients. All images acquired from the three different sources: (1) The Center for Biomedical Image Computing and Analytics (CBICA) (2) The Cancer Imaging Archive (TCIA) data center (3) The data from other sites(Other). 
We use 210 HGG cases in the challenge training set in our study since we have no access to the test data of the BraTS2018 Challenge. We split the 210 cases into train (170 cases) and test (40 cases) sets. 
Each case has four types of MRI scans (T1, T1c, T2 and FLAIR) and three types of tumor sub-region labels. All modalities have been aligned to a common space and resampled to 1mm isotropic resolution~\cite{bakas2018identifying}.

In our experiments, we evaluate our method to learn the distribution of all HGG cases across different data centers. In the GAN synthesis phase, all three labels are utilized to generate fake images. For segmentation, we focus on the whole tumor region (union of all three labels). The image dataset used in each experiment share one or multiple modalities.
Without loss of generality, we picked T2, T1+T2+FLAIR, T1c+T2+FLAIR modalities respectively for the following three experiments.


\subsubsection{Evaluation metrics}
The Dice score (Dice), sensitivity (Sens), specificity (Spec), and 95\% quantile of Hausdorff distance (HD95) are adopted to evaluate the segmentation performance of brain tumor, same as the metrics in the BraTS2018 Challenge~\cite{bakas2018identifying}. 
The Dice score, sensitivity (true positive rate) and specificity (true negative rate) measure the overlap between ground-truth mask $G$ and segmented result $S$. They are defined as
\begin{equation}
Dice(G, S) = \frac{2|G \cap S|}{|G| + |S|}
\end{equation}
\begin{equation}
Sens(G, S)=\frac{|G \cap S|}{|G|}
\end{equation}
\begin{equation}
Spec(G, S)=\frac{|(1-G) \cap (1-S)|}{|1-G|}
\end{equation}
The Hausdorff distance evaluates the distance between boundaries of ground-truth and segmented masks:
\begin{equation}
HD(G, S) = \max\{\sup_{x\in\partial G}\inf_{y\in\partial S}d(x, y), \sup_{y\in\partial S}\inf_{x\in\partial G}d(x, y)\}
\end{equation}
where $\partial$ means the boundary operation, and $d$ is Euclidean distance. Because the Hausdorff distance is sensitive to small outlying subregions, we use the 95\% quantile of the distances (named as HD95). We calculate the 3D metrics of each studies for all the experiments and prevent the small object exclusion described in \cite{chang2020synthetic}.


\subsection{Implementation details}
In the training generator phase, we use 9-blocks ResNet~\cite{he2016deep} architecture for the generator and multiple discriminators which have the same structure as that in PatchGAN~\cite{pix2pix2017} with patch size $70\times70$. We resize the input image as $286\times286$ and then randomly crop the image to $256\times256$. We apply the Adam optimizer~\cite{kingma2014adam} in our experiments, with a learning rate of 0.0001, and momentum parameters $\beta_1 = 0.5$, $\beta_2 = 0.999$.
We use batch size $m=10$ for all BraTS2018 dataset. 
Our training time is approximately two days with one 48G memory Nvidia Quadro RTX 8000 GPU.

In the segmentation phase, we resize the input image as $286\times286$ and then randomly crop the image to $256\times256$ with a batch size of 16 as input. The model is trained with Adam optimizer using a learning rate of 0.001 for 50 epochs in brain tumor segmentation. 
We use random data augmentation in all experiments, including random flip and random scales to improve performance. During testing, we predict segmentation for each 2D slice and then calculate the evaluation metrics on 3D volume.

\subsection{Experiment on homogeneous and single-modal datasets}\label{sec:singlemod}
In this section, we show that the proposed AsynDGAN can learn the distributions and generate realistic synthetic medical images across all the homogeneous data centers. In other words, the downstream task (here we use segmentation)'s model that learned from the synthetic images could outperform the model that learned from the single data center's real images and achieve comparable results to the model learned from all the real images.

\begin{table}[t]
    \caption{\label{tab:hgg-10}Brain tumor segmentation results using the homogeneous and single modality(T2) dataset. The Real-Subset-n are randomly split the whole training set equally.}
	\begin{center}
		\begin{tabular}{p{0.11\textwidth}p{0.06\textwidth}p{0.06\textwidth}p{0.06\textwidth}p{0.06\textwidth}}
			\toprule
			Method & Dice(\%) $\uparrow$ & Sens(\%) $\uparrow$  & Spec(\%) $\uparrow$  & HD95 $\downarrow$ \\
			\midrule
			Real-All & 80.8\textpm16.2 & 78.5\textpm2.14	& 99.6\textpm0.2 & 11.95\textpm7.6 \\ \midrule
			Real-Subset-1 & 75.2\textpm19.1 &    72.4\textpm23.1 &    99.5\textpm0.5 &    18.56\textpm11.0 \\
			Real-Subset-2 & 71.4\textpm23.5 &	67.3\textpm27.5 &	99.6\textpm0.5 &	17.66\textpm10.4 \\
			Real-Subset-3 & 72.0\textpm23.5 &	67.1\textpm26.2 &	99.7\textpm0.3 &	19.99\textpm13.8 \\
			Real-Subset-4 &71.4\textpm22.6 &	66.6\textpm25.9 &	99.6\textpm0.3 &	18.37\textpm10.9 \\
			Real-Subset-5 & 75.5\textpm21.3 & 71.6\textpm24.2 & 99.6\textpm0.2 & 15.03\textpm9.23 \\
			Real-Subset-6 & 70.0\textpm20.7 & 64.0\textpm22.4 & 99.5\textpm0.7 & 19.33\textpm11.2 \\
			Real-Subset-7 & 74.5\textpm18.8 & 68.4\textpm22.6 & 99.7\textpm0.2 & 19.89\textpm9.08 \\
			Real-Subset-8 & 72.7\textpm21.2 & 68.5\textpm24.5 & 99.5\textpm0.4 & 18.47\textpm10.8 \\
			Real-Subset-9 & 73.1\textpm21.5 & 71.4\textpm26.3 & 99.5\textpm0.4 & 16.06\textpm8.43 \\
			Real-Subset-10 & 72.1\textpm23.1 & 68.4\textpm26.0 & 99.6\textpm0.3 & 20.39\textpm13.9 \\ \midrule
			\textbf{AsynDGAN}  & 77.3\textpm18.0 & 74.2\textpm22.3 & 99.6\textpm0.3 & 16.44\textpm11.6 \\
			\bottomrule
		\end{tabular}
	\end{center}
\end{table}



\subsubsection{Settings}
The training set, which is randomly divided into 10 subsets equally (17 cases each) as the homogeneous subset, is treated as 10 distributed medical entities. There are a total of 170 cases (11,349 images) in the training set and 40 cases (2,730 images) in the test set.
 We conduct the following segmentation experiments:
(1) Real-All. Training using real images from the whole training set (170 cases).
(2) Real-Subset-n. Training using real images from the $n$-th subset (medical entity), where $n=1,2,\cdots,10$. The datasets are randomly split the whole training set equally. 
There are 10 different experiments in this category.
(3) AsynDGAN. Training using synthetic images from our proposed AsynDGAN.
The AsynDGAN is trained using images from the 10 subsets (medical entities) in a distributed fashion.

In all experiments, the test set remains the same for a fair comparison. It should be noted that in the AsynDGAN experiments, the number of synthetic images is the same as that of real images in Real-All. The regular GAN has the same generator and discriminator structures as AsynDGAN, as well as the hyper-parameters. The key difference is that AsynDGAN has 10 different discriminators, and each of them is located in one medical entity and only has access to the real images in one subset. It should also be noted that the setting is improved from \cite{chang2020synthetic}, where the dataset is split by tumor size.

\begin{figure}[t]
	\begin{center}
	    \includegraphics[width=0.3\linewidth]{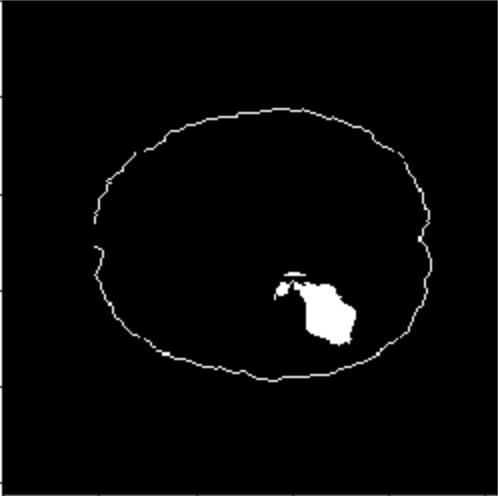}
		\includegraphics[width=0.3\linewidth]{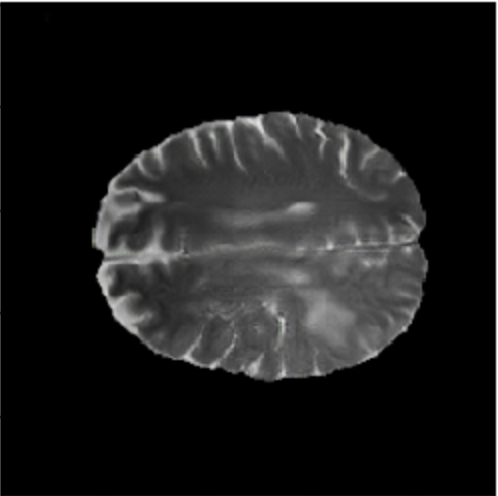}
		\includegraphics[width=0.3\linewidth]{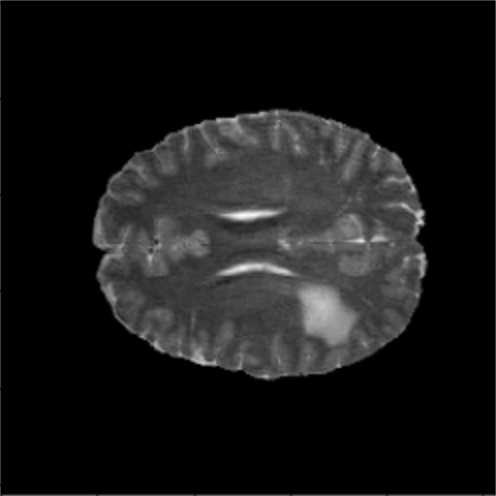}\\ \vspace{0.01in}
		\begin{minipage}{0.3\linewidth}
			\centering\includegraphics[width=\linewidth]{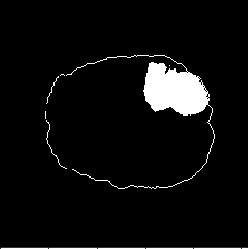} \\ (a) Input
		\end{minipage}
		\begin{minipage}{0.3\linewidth}
			\centering\includegraphics[width=\linewidth]{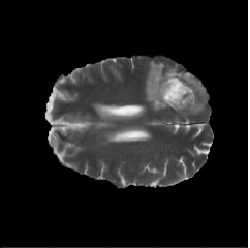} \\ (b) AsynDGAN
		\end{minipage}
		\begin{minipage}{0.3\linewidth}
			\centering\includegraphics[width=\linewidth]{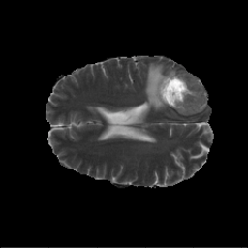}  \\  (c) Real
		\end{minipage}
	\end{center}
	\caption{The examples of synthetic brain tumor images from the AsynDGAN. (a) The input of the AsynDGAN network. (b) Synthetic single modality images of AsynDGAN based on the input. (c) Real single modality images.}
	\label{fig:syn:hgg-10}
\end{figure}

\begin{figure*}[t]
	\begin{center}
		\includegraphics[width=0.15\linewidth]{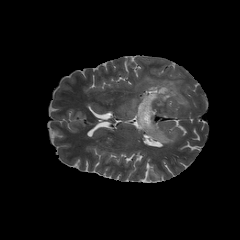}
		\includegraphics[width=0.15\linewidth]{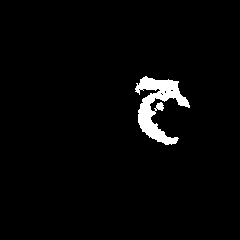}
		\includegraphics[width=0.15\linewidth]{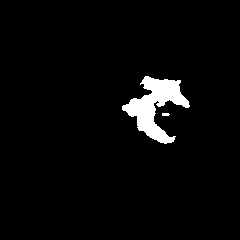}
		\includegraphics[width=0.15\linewidth]{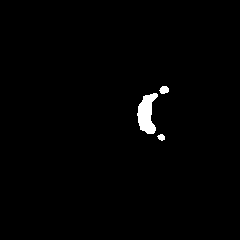}
		\includegraphics[width=0.15\linewidth]{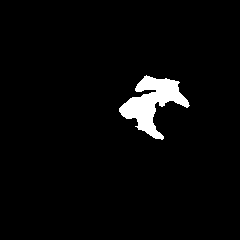} \\
		\vspace{0.01in}
		\begin{minipage}{0.15\linewidth}
			\centering\includegraphics[width=\linewidth]{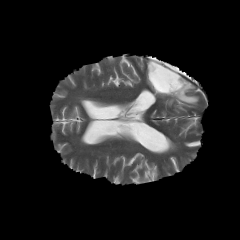} \\ (a) Image
		\end{minipage}
		\begin{minipage}{0.15\linewidth}
			\centering\includegraphics[width=\linewidth]{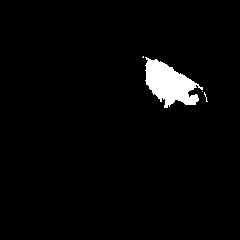} \\ (b) Label
		\end{minipage}
		\begin{minipage}{0.15\linewidth}
			\centering\includegraphics[width=\linewidth]{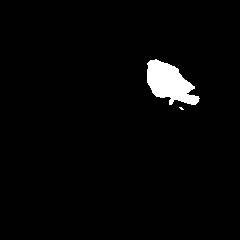}  \\  (c) Real-All
		\end{minipage}
		\begin{minipage}{0.15\linewidth}
			\centering\includegraphics[width=\linewidth]{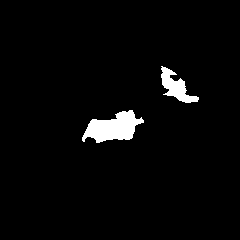} \\ (e) Real-Subset-6
		\end{minipage}
		\begin{minipage}{0.15\linewidth}
			\centering\includegraphics[width=\linewidth]{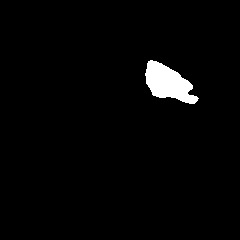} \\ (f) AsynDGAN
		\end{minipage}
	\end{center}
	\caption{Typical brain tumor segmentation results. (a) Test images. (b) Ground-truth labels of tumor region. (c)-(f) are results of models trained on all real images, synthetic images of regular GAN, real images from subset-6, synthetic images of AsynDGAN, respectively.}
	\label{fig:seg:hgg-10}
\end{figure*}

\subsubsection{Results}
The quantitative brain tumor segmentation results are shown in Table~\ref{tab:hgg-10}. The model trained using all real images (Real-All) is the ideal case assuming that we can access all data. It is our baseline and achieves the best performance. Compared with the ideal baseline, the performance of models trained using data in each medical entity (Real-Subset-1$\sim$10) degrades a lot, because the information in each subset is limited and the number of training images is much smaller.

Our AsynDGAN can learn from the information of all data during training, although the generator doesn't ``see`` the real images. And we can generate as many synthetic images as we want to train the segmentation model. Therefore, the model (AsynDGAN) outperforms all models using a single subset. 
The examples of synthetic images from AysnDGAN are shown in Fig~\ref{fig:syn:hgg-10}. Several qualitative segmentation results of each method are shown in Fig~\ref{fig:seg:hgg-10}. It should be noted that the split strategy differences lead our results are better than the experiments in \cite{chang2020synthetic}, which split the dataset by the tumor size in the training set but test on the randomly sampled dataset.

\subsection{Experiment on heterogeneous and multi-modal datasets}

\subsubsection{Settings}
In this section, we show that our AsynDGAN can learn the distributions and generate realistic multi-modality medical images across all the heterogeneous data centers.
Specifically, the generator could generate realistic three channels (T1, T2, and Flair) multi-modality images by learning from three heterogeneous data sources.

The training data is split into 3 subsets based on the different sources of the data described in \cite{menze2014multimodal}:
(1) Real-CBICA, data collected from the Center for Biomedical Image Computing and Analytics(CBICA). In total, we have 88 studies from this subset.
(2) Real-TCIA, data collected from the Cancer Imaging Archive(TCIA). In total, we have 102 studies from this subset.
(3) Real-Other, the data collected not from CBICA nor TCIA. In total, we have 20 studies from this subset.

Besides, we conduct additional experiments to evaluate the performance of the segmentation model by learning from the synthetic images jointly with one of the real-image subsets:
(1) Syn+Real-CBICA includes the synthetic data generated from AsynDGAN combined with the real images in the CBICA subset. (2) Syn+Real-TCIA includes the synthetic data and the real images in the TCIA subset.
(3) Syn+Real-Other includes the synthetic data and the real images in the other subset. 
In this way, we treat the synthetic images as a novel data augmentation method~\cite{antoniou2017data}. 

\subsubsection{Result}

The multi-modality three data centers brain tumor segmentation results are shown in Table~\ref{tab:hgg-3}. 
Same as the single modality experiments, the model trained using all real images (Real-All) is the ideal case that we can access all data. It is our baseline and achieves the best performance. Compared with the ideal baseline, the performance of models trained only uses data in each medical entity (Real-CBICA, Real-TCIA, Real-Other) degrades a lot. 

Unlike the single modality experiments, the dataset divided unevenly. 
Our method can learn evenly from the information of all datasets during training, although the generator doesn't ``see`` the real images. Therefore, the model (AsynDGAN) outperforms all models learn from a single subset.

\begin{figure*}[t]
	\begin{center}

	    \includegraphics[width=0.12\linewidth]{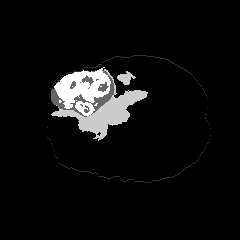}
	    \includegraphics[width=0.12\linewidth]{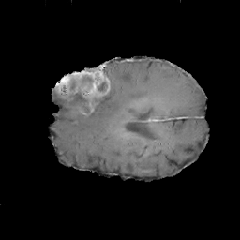}
	    \includegraphics[width=0.12\linewidth]{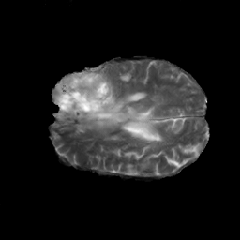}
	    \includegraphics[width=0.12\linewidth]{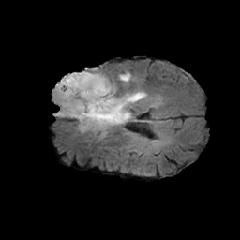}
	    \includegraphics[width=0.12\linewidth]{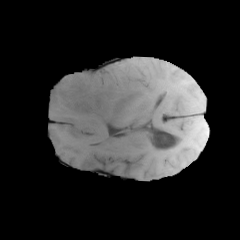}
	    \includegraphics[width=0.12\linewidth]{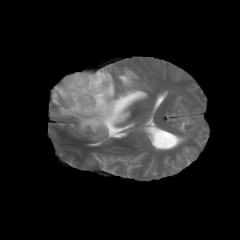}
	    \includegraphics[width=0.12\linewidth]{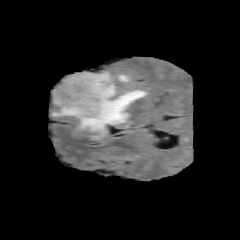}\\ 
	    \vspace{0.01in}
	    \includegraphics[width=0.12\linewidth]{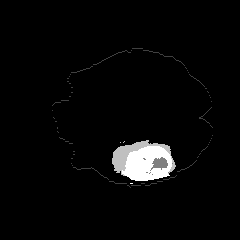}
		\includegraphics[width=0.12\linewidth]{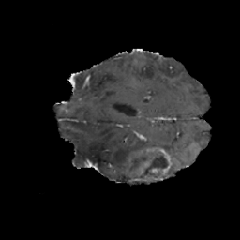}
		\includegraphics[width=0.12\linewidth]{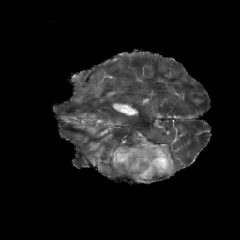}
	    \includegraphics[width=0.12\linewidth]{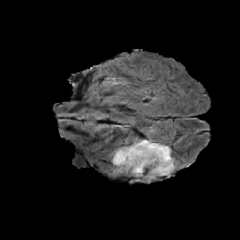}
		\includegraphics[width=0.12\linewidth]{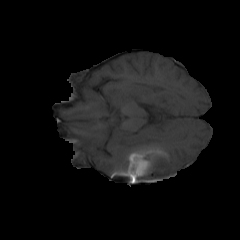}
		\includegraphics[width=0.12\linewidth]{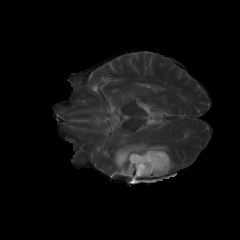}
		\includegraphics[width=0.12\linewidth]{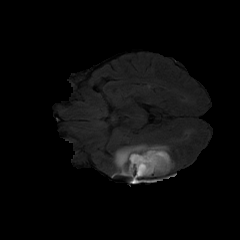}\\ 
		\vspace{0.01in}
		\begin{minipage}{0.12\linewidth}
	    \includegraphics[width=\linewidth]{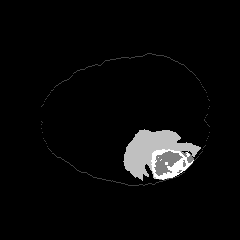}\\(a)Input
	    \end{minipage}
	    \begin{minipage}{0.12\linewidth}
	    \includegraphics[width=\linewidth]{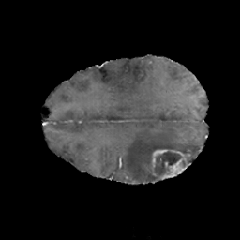}\\(b) Syn T1
	    \end{minipage}
	    \begin{minipage}{0.12\linewidth}
	    \includegraphics[width=\linewidth]{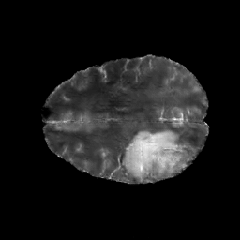}\\(c) Syn T2
	    \end{minipage}
	    \begin{minipage}{0.12\linewidth}
	    \includegraphics[width=\linewidth]{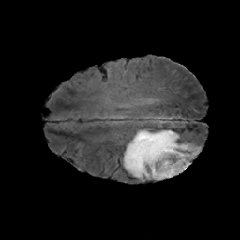}\\(d) Syn Flair
	    \end{minipage}
	    \begin{minipage}{0.12\linewidth}
	    \includegraphics[width=\linewidth]{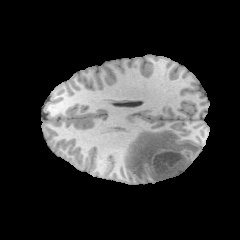}\\(e) Real T1
	    \end{minipage}
	    \begin{minipage}{0.12\linewidth}
	    \includegraphics[width=\linewidth]{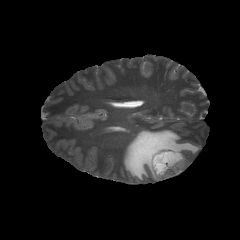}\\(f) Real T2
	    \end{minipage}
	    \begin{minipage}{0.12\linewidth}
	    \includegraphics[width=\linewidth]{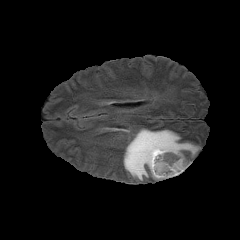}\\(g) Real Flair
	    \end{minipage}
	\end{center}
	\caption{The examples of multi-modality synthetic brain tumor images from the AsynDGAN. (a) The input of the AsynDGAN. (b)-(d) Synthetic multi-modality images of AsynDGAN. (e)-(g) Real multi-modality images.}
	\label{fig:syn:hgg-multimod}
\end{figure*}

We can see from the last three rows of Table~\ref{tab:hgg-3} that they achieve higher performance compared with the corresponding real-image only subsets and are even better than the Real-All result. That means the synthetic images generated by our framework can not only be used as a stand-alone database in down-stream segmentation tasks but also an efficient data augmentation to a real-image dataset to boost the performance. 
Some examples of synthetic images and corresponding real images are shown in Fig.~\ref{fig:syn:hgg-multimod}

\begin{table}[t]
    \caption{\label{tab:hgg-3}Brain tumor segmentation results over three heterogeneous and multi-modal (T1+T2+Flair) subsets.}
	\begin{center}
		\begin{tabular}{p{0.134\textwidth}p{0.05\textwidth}p{0.05\textwidth}p{0.05\textwidth}p{0.06\textwidth}}
			\toprule
			Method & Dice(\%) $\uparrow$ & Sens(\%) $\uparrow$  & Spec(\%)$\uparrow$  & HD95$\downarrow$ \\
			\midrule
			Real-All & 85.9\textpm12.7 & 82.3\textpm17.8 & 99.7\textpm0.2 & 8.970\textpm5.11 \\ \midrule
			Real-CBICA & 78.9\textpm19.6 & 75.7\textpm23.1 & 99.7\textpm0.2 & 16.45\textpm9.89 \\
			Real-TCIA & 77.2\textpm12.1 & 82.1\textpm16.1 & 99.3\textpm0.4 & 12.68\textpm4.95 \\
			Real-Other & 80.4\textpm12.9 & 80.7\textpm19.4 & 99.5\textpm0.3 & 23.33\textpm14.0 \\ \midrule
			\textbf{AsynDGAN}  & 82.0\textpm17.6 & 81.9\textpm22.0 & 99.5\textpm0.6 & 13.93\textpm10.0 \\ \midrule
			\textbf{Syn + Real-CBICA} & 85.3\textpm13.9 & 83.3\textpm18.7 & 99.6\textpm0.2 & 10.95\textpm7.30\\
			\textbf{Syn + Real-TCIA} & 86.9\textpm9.7 & 88.3\textpm15.1 & 99.5\textpm0.3 & 11.08\textpm6.15 \\
			\textbf{Syn + Real-Other} & 84.7\textpm15.3 & 83.1\textpm18.6 & 99.7\textpm0.2 & 12.87\textpm10.4 \\
			\bottomrule
		\end{tabular}
	\end{center}
\end{table}

\subsection{Experiment on missing-modality datasets}
In this section, we train the AsynDGAN with the following missing modality datasets:

    (1) Real-CBICA(n/a:T2) subset: real T1c, Flair, and missing T2.
    
    (2) Real-TCIA(n/a:Flair) subset: real T1c, T2, and missing Flair.
    
    (3) Real-Other(n/a:T1c) subset: real T2, Flair, and missing T1c.
    
Similar as the previous section, AsynDGAN still generates three modalities at the same time, but this time it only collects the losses for the real image channels.

\subsubsection{Settings}
In this section, we demonstrate that the proposed AsynDGAN could adapt the modality variances across data centers. 
In each data center, we exclude one of these three modalities from three data centers. 
Although each dataset has a missing modality, we show that AsynDGAN can still learn to generate all modalities from available data across all centers.

One of the advantages of learning to generate multiple modalities by using our framework is that the learned generator can be used for random missing modality completion. For instance, with a well-trained $G$, we can synthesize the T2 images for Real-CBICA(n/a:T2) subset so that the center could utilize all information from different modalities in their study.
To evaluate the effect of missing modality completion by the synthetic images, we conduct the following segmentation experiments after learning AsynDGAN with missing modality datasets: 
(1) Real-CBICA(n/a:T2), Real-TCIA(n/a:Flair), Real-Other(n/a:T1c). Training segmentation networks by using real data from different medical centers with various missing modalities.
(2)Completed-CBICA (syn:T2): Real-CBICA's real T1c, Flair combined with synthetic T2 generated by AsynDGAN. Same study size as Real-CBICA.
(3)Completed-TCIA (syn:Flair): Real-TCIA's real T1c and T2 combined with synthetic Flair generated by AsynDGAN. Same study size as Real-TCIA.
(4)Completed-Other (syn:T1c): Real-Other's real T2 and Flair combined with synthetic T1c generated by AsynDGAN. Same study size as Real-Other.

\subsubsection{Result}

The brain tumor segmentation results about modality completion are shown in Table \ref{tab:hgg-4}. Overall, the performances of the segmentation models trained on three data entities (Real-CBICA(n/a:T2), Real-TCIA(n/a:Flair), Real-Other(n/a:T1c)) with missing modality are worse than the models trained on the completed datasets with combination of real and synthetic images. 
Specifically, by combining synthetic T2 with the real T1c and Flair images, the model trained on Completed-CBICA (syn:T2) can achieve superior performance to the model trained on  Real-CBICA(n/a:T2) subset. We have similar observations for the Completed-TCIA (syn:Flair). 
Besides, we also notice that T2 and Flair may contribute more to the whole tumor segmentation task since learning from the smallest subset Real-Other(n/a:T1c) achieves higher performance compared with learning from the other subsets with missing T2 or Flair. As a result, there is no significant difference between Completed-Other (syn:T1c) and Real-Other(n/a:T1c) by introducing the synthetic T1c images.

\begin{figure*}[t]
	\begin{center}
		\includegraphics[width=\linewidth]{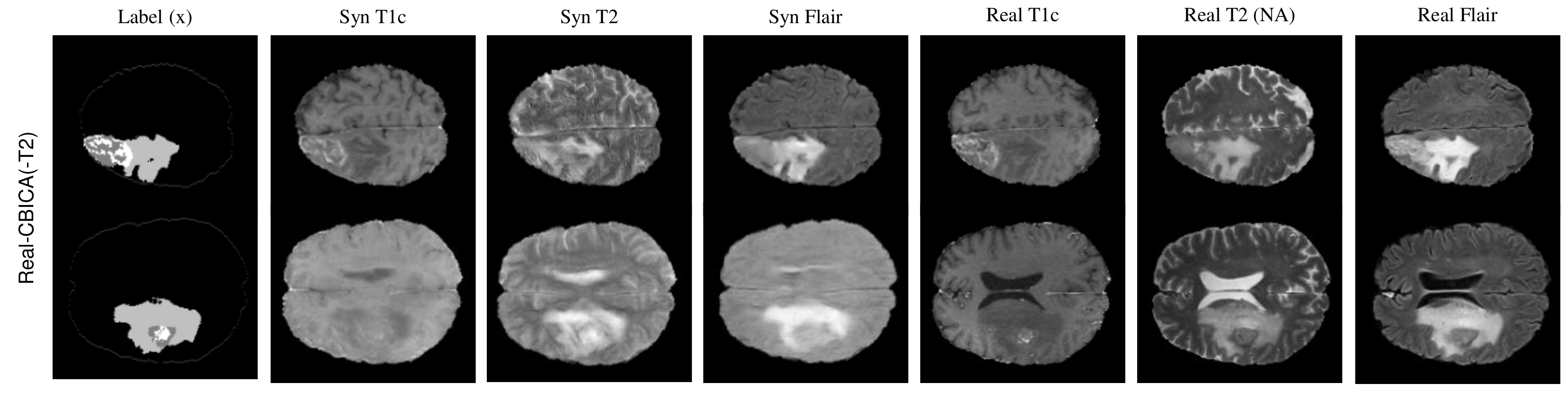}
		\includegraphics[width=\linewidth]{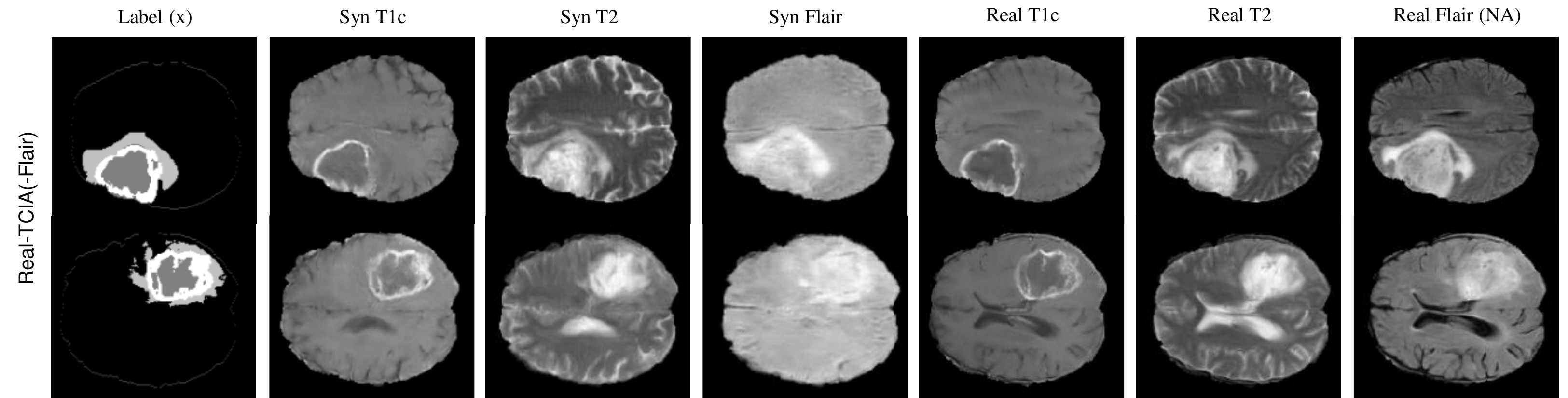}
		\includegraphics[width=\linewidth]{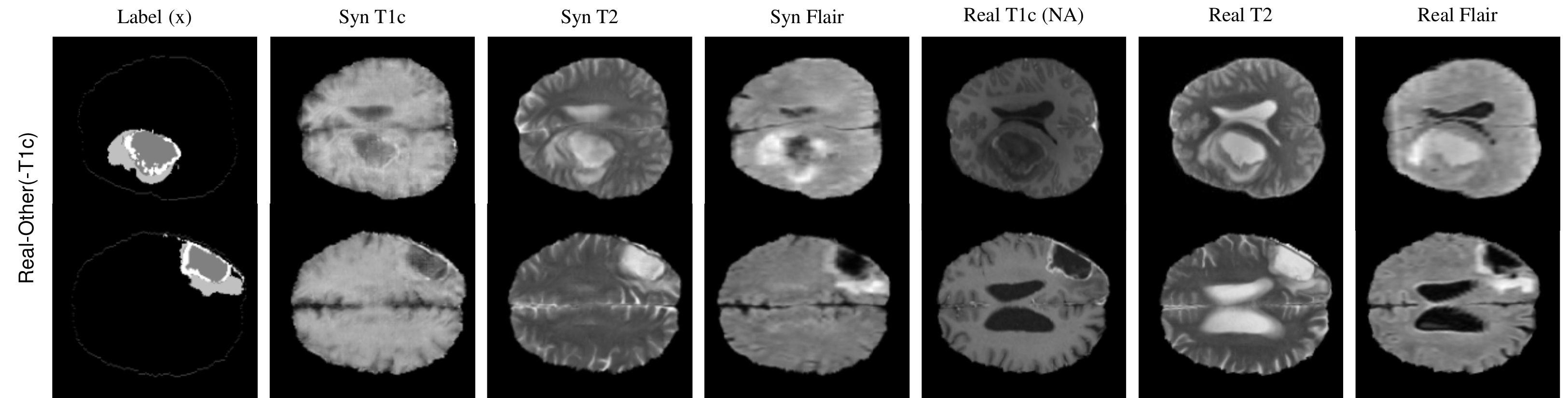}
	\end{center}
	\caption{The examples of synthetic brain tumor images by the AsynDGAN after learning from multiple missing-modality datasets. `NA' column indicates the missing modality during training.}
	\label{fig:syn:missing}
\end{figure*}

We show some examples of synthetic images and corresponding real images in Fig.~\ref{fig:syn:missing}. In this figure, 3 sections are corresponding to three data centers, respectively. The column of the real image labeled as NA (not available) indicates the missing modality in that center during the training of AsynDGAN. The first observation is that our method can still learn to generate multiple modalities when centers have missing modality. We also notice that the synthetic images may not have the same global context as the real images, for example, the generated brains may have different shapes of ventricles. This is due to the lack of information about other tissues outside the tumor region in the input of the $G$.
On one hand, this variation is good for privacy preservation. On the other hand, for missing modality completion, the synthetic modality may have a different context with the real modalities.
However, this limitation seems not critical in our segmentation task, since the results in Table~\ref{tab:hgg-4} show clear improvement after the missing modality completion.

\begin{table}[t]
    \caption{\label{tab:hgg-4}Brain tumor segmentation results over three datasets with missing modality (T1c/T2/Flair)}
	\begin{center}
		\begin{tabular}{p{0.15\textwidth}p{0.05\textwidth}p{0.05\textwidth}p{0.05\textwidth}p{0.06\textwidth}}
			\toprule
			Method & Dice(\%)~$\uparrow$ & Sens(\%)~$\uparrow$  & Spec(\%)~$\uparrow$  & HD95~$\downarrow$ \\
			\midrule
			Real-CBICA(n/a:T2) & 78.0\textpm23.4 & 74.5\textpm25.9 & 99.7\textpm0.2 & 15.47\textpm14.2 \\
			Real-TCIA(n/a:Flair) & 76.7\textpm15.3 & 72.8\textpm20.8 & 99.5\textpm0.8 & 15.64\textpm8.75 \\
			Real-Other(n/a:T1c) & 80.9\textpm14.1 & 79.3\textpm18.8 & 99.6\textpm0.2 & 16.74\textpm9.41\\ \midrule
			
			\textbf{Completed-CBICA (syn:T2)}  & 83.0\textpm14.6 & 79.9\textpm19.0 & 99.7\textpm0.2 & 15.64\textpm9.93\\
			\textbf{Completed-TCIA (syn:Flair)}& 85.5\textpm10.4 & 83.3\textpm14.3 & 99.7\textpm0.1 & 15.02\textpm8.35  \\
            \textbf{Completed-Other (syn:T1c)} & 80.9\textpm15.3 & 80.6\textpm19.1 & 99.6\textpm0.2 & 16.93\textpm11.8 \\
			\bottomrule
		\end{tabular}
	\end{center}
\end{table}

